\documentclass{article}

\usepackage{amsmath,amsfonts,amssymb,times,graphicx,natbib,algorithm,algorithmic,hyperref}

\usepackage[accepted]{whi2016}

\icmltitlerunning{Are RNNs and HMMs More Interpretable When Combined?}

\graphicspath{ {./graphs/}}
\usepackage{subcaption}
\usepackage{multirow}

\begin{document} 

\twocolumn[
\icmltitle{Increasing the Interpretability of Recurrent Neural Networks\\Using Hidden Markov Models}
\icmlauthor{Viktoriya Krakovna}{vkrakovna@fas.harvard.edu}
\icmladdress{Department of Statistics, Harvard University}
\icmlauthor{Finale Doshi-Velez}{finale@seas.harvard.edu}
\icmladdress{Department of Computer Science, Harvard University}

\vskip 0.3in
]

\begin{abstract}
As deep neural networks continue to revolutionize various application domains, there is increasing interest in making these powerful models more understandable and interpretable, and narrowing down the causes of good and bad predictions. We focus on recurrent neural networks (RNNs), state of the art models in speech recognition and translation. Our approach to increasing interpretability is by combining an RNN with a hidden Markov model (HMM), a simpler and more transparent model.  We explore various combinations of RNNs and HMMs: an HMM trained on LSTM states; a hybrid model where an HMM is trained first, then a small LSTM is given HMM state distributions and trained to fill in gaps in the HMM's performance; and a jointly trained hybrid model. We find that the LSTM and HMM learn complementary information about the features in the text.
\end{abstract}

\section{Introduction}
Following the recent progress in deep learning, researchers and practitioners of machine learning are recognizing the importance of understanding and interpreting what goes on inside these black box models. Recurrent neural networks have recently revolutionized speech recognition and translation, and these powerful models could be very useful in other applications involving sequential data. However, adoption has been slow in applications such as health care, where practitioners are reluctant to let an opaque expert system make crucial decisions. If we can make the inner workings of RNNs more interpretable, more applications can benefit from their power.

There are several aspects of what makes a model or algorithm understandable to humans. One aspect is model complexity or parsimony. Another aspect is the ability to trace back from a prediction or model component to particularly influential features in the data \cite{Ruping}  \cite{Kim}. This could be useful for understanding mistakes made by neural networks, which have human-level performance most of the time, but can perform very poorly on seemingly easy cases. For instance, convolutional networks can misclassify adversarial examples with very high confidence \citep{Nguyen}, and made headlines in 2015 when the image tagging algorithm in Google Photos mislabeled African Americans as gorillas. It's reasonable to expect recurrent networks to fail in similar ways as well. It would thus be useful to have more visibility into where these sorts of errors come from, i.e. which groups of features contribute to such flawed predictions. 

Several promising approaches to interpreting RNNs have been developed recently. \citet{Che} have approached this by using gradient boosting trees to predict LSTM output probabilities and explain which features played a part in the prediction. They do not model the internal structure of the LSTM, but instead approximate the entire architecture as a black box. \citet{Karpathy} showed that in LSTM language models, around 10\% of the memory state dimensions can be interpreted with the naked eye by color-coding the text data with the state values; some of them track quotes, brackets and other clearly identifiable aspects of the text. Building on these results, we take a somewhat more systematic approach to looking for interpretable hidden state dimensions, by using decision trees to predict individual hidden state dimensions (Figure \ref{fig:decision_trees}). We visualize the overall dynamics of the hidden states by coloring the training data with the k-means clusters on the state vectors (Figures \ref{subfig:shak_lstm}, \ref{subfig:linux_lstm}). 

We explore several methods for building interpretable models by combining LSTMs and HMMs. The existing body of literature mostly focuses on methods that specifically train the RNN to predict HMM states \citep{Bourlard} or posteriors \citep{Maas}, referred to as hybrid or tandem methods respectively.
We first investigate an approach that does not require the RNN to be modified in order to make it understandable, as the interpretation happens after the fact. Here, we model the big picture of the state changes in the LSTM, by extracting the hidden states and approximating them with a continuous emission hidden Markov model (HMM). We then take the reverse approach where the HMM state probabilities are added to the output layer of the LSTM (see Figure \ref{fig:flowchart}). The LSTM model can then make use of the information from the HMM, and fill in the gaps when the HMM is not performing well, resulting in an LSTM with a smaller number of hidden state dimensions that could be interpreted individually (Figures \ref{fig:viz_hybrid_shak}, \ref{fig:viz_hybrid_linux}). 

\section{Methods}
We compare a hybrid HMM-LSTM approach with a continuous emission HMM (trained on the hidden states of a 2-layer LSTM), and a discrete emission HMM (trained directly on data).

\subsection{LSTM models} 

We use a character-level LSTM with 1 layer and no dropout, based on the Element-Research library. We train the LSTM for 10 epochs, starting with a learning rate of 1, where the learning rate is halved whenever $\exp(-l_t) > \exp(-l_{t-1}) + 1$, where $l_t$ is the log likelihood score at epoch $t$. The $L_2$-norm of the parameter gradient vector is clipped at a threshold of 5.

%\textbf{FINALE: GIVE BASIC INFO/PARAMETERS} 

\begin{figure}[t]
\centering
\includegraphics[scale=.35]{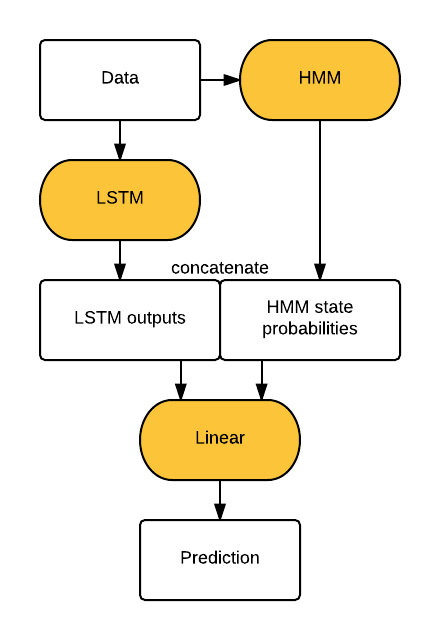}
\caption{Hybrid HMM-LSTM algorithm.}
\label{fig:flowchart}
\end{figure}

\subsection{Hidden Markov models}

%\textbf{FINALE: CAN TRIM INFERENCE IF NEEDED} 

The HMM training procedure is as follows:

\textbf{Initialization of HMM hidden states:}
\begin{itemize}
\item[] (Discrete HMM) Random multinomial draw for each time step (i.i.d. across time steps).
\item[] (Continuous HMM) K-means clusters fit on LSTM states, to speed up convergence relative to random initialization. 
\end{itemize}

\textbf{At each iteration:}
\begin{enumerate}
	\item Sample states using Forward Filtering Backwards Sampling algorithm (FFBS, \citet{Rao}).
	\item Sample transition parameters from a Multinomial-Dirichlet posterior.
	Let $n_{ij}$ be the number of transitions from state $i$ to state $j$. Then the posterior distribution of the $i$-th row of transition matrix $T$ (corresponding to transitions from state $i$) is: 
	$$T_i \sim \text{Mult}(n_{ij} | T_i) \text{Dir}(T_i | \alpha)$$
	where $\alpha$ is the Dirichlet hyperparameter.
	\item (Continuous HMM) Sample multivariate normal emission parameters from Normal-Inverse-Wishart posterior for state $i$: 
	
    $$ \mu_i, \Sigma_i \sim N(y|\mu_i, \Sigma_i) N(\mu_i |0, \Sigma_i) \text{IW}(\Sigma_i) $$
    
    (Discrete HMM) Sample the emission parameters from a Multinomial-Dirichlet posterior.
\end{enumerate}

\textbf{Evaluation:}

We evaluate the methods on how well they predict the next observation in the validation set. For the HMM models, we do a forward pass on the validation set (no backward pass unlike the full FFBS), and compute the HMM state distribution vector $p_t$ for each time step $t$. Then we compute the predictive likelihood for the next observation as follows:
$$ P(y_{t+1} | p_t) =\sum_{x_t=1}^n \sum_{x_{t+1}=1}^n p_{tx_t} \cdot T_{x_t, x_{t+1}} \cdot P(y_{t+1} | x_{t+1})$$

where $n$ is the number of hidden states in the HMM. 

\begin{figure*}[t]
\centering
\includegraphics[scale=.5]{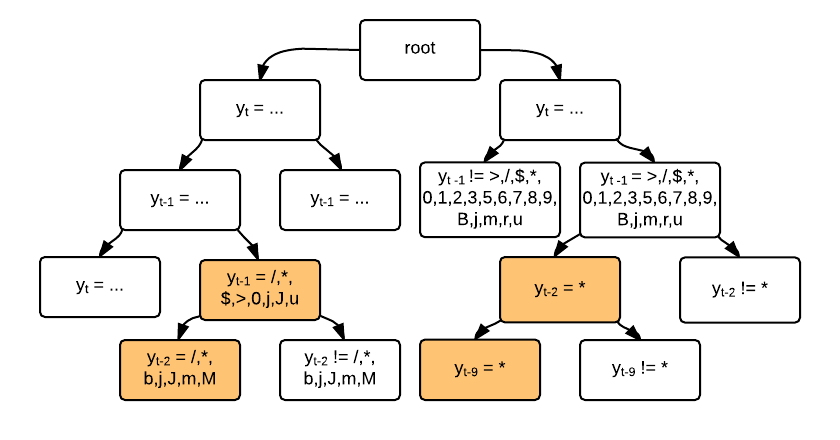}
\caption{Decision tree predicting an individual hidden state dimension of the hybrid algorithm based on the preceding characters on the Linux data. The hidden state dimensions of the 10-state hybrid mostly track comment characters.}
\label{fig:decision_trees}
\end{figure*}

\subsection{Hybrid models}

Our main hybrid model is put together sequentially, as shown in Figure \ref{fig:flowchart}. We first run the discrete HMM on the data, outputting the hidden state distributions obtained by the HMM's forward pass, and then add this information to the architecture in parallel with a 1-layer LSTM. The linear layer between the LSTM and the prediction layer is augmented with an extra column for each HMM state. The LSTM component of this architecture can be smaller than a standalone LSTM, since it only needs to fill in the gaps in the HMM's predictions. The HMM is written in Python, and the rest of the architecture is in Torch.

We also build a joint hybrid model, where the LSTM and HMM are simultaneously trained in Torch. We implemented an HMM Torch module, optimized using stochastic gradient descent rather than FFBS. Similarly to the sequential hybrid model, we concatenate the LSTM outputs with the HMM state probabilities.  

%\textbf{FINALE: TO ADD TO TABLE, CAN YOU SEE WHAT DEPTH DECISION TREE IS NEEDED FOR THE CELL STATES, ON AVERAGE, FOR EACH LSTM? OR AN ERROR MEASURE OF A FIXED DEPTH? TO SEE IF THE STATES ARE SOMEHOW SIMPLER?} 

\section{Experiments}
We test the models on several text data sets on the character level: the Penn Tree Bank (5M characters), and two data sets used by \citet{Karpathy}, Tiny Shakespeare (1M characters) and Linux Kernel (5M characters).  We chose $k=20$ for the continuous HMM based on a PCA analysis of the LSTM states, as the first 20 components captured almost all the variance. 

Table \ref{tab:results} shows the predictive log likelihood of the next text character for each method. On all text data sets, the hybrid algorithm performs a bit better than the standalone LSTM with the same LSTM state dimension. This effect gets smaller as we increase the LSTM size and the HMM makes less difference to the prediction (though it can still make a difference in terms of interpretability). The hybrid algorithm with 20 HMM states does better than the one with 10 HMM states. The joint hybrid algorithm outperforms the sequential hybrid on Shakespeare data, but does worse on PTB and Linux data, which suggests that the joint hybrid is more helpful for smaller data sets. The joint hybrid is an order of magnitude slower than the sequential hybrid, as the SGD-based HMM is slower to train than the FFBS-based HMM.

%\textbf{FINALE: NEEDS MORE DISCUSSION ON WHAT YOU THINK IS GOING ON} 

\begin{figure*}
\centering
\begin{subfigure}{0.49\textwidth}
\centering
\includegraphics[scale=.73]{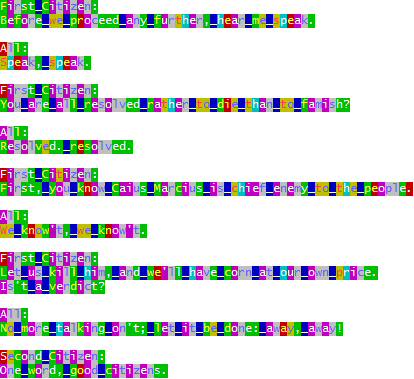}
\caption{Hybrid HMM component: colors correspond to 10 HMM states. Blue cluster identifies spaces. Green cluster (with white font) identifies punctuation and ends of words. Purple cluster picks up on some vowels.}
\label{subfig:shak_hmm}
\end{subfigure}\hfill
\begin{subfigure}{0.49\textwidth}
\centering
\includegraphics[scale=.73]{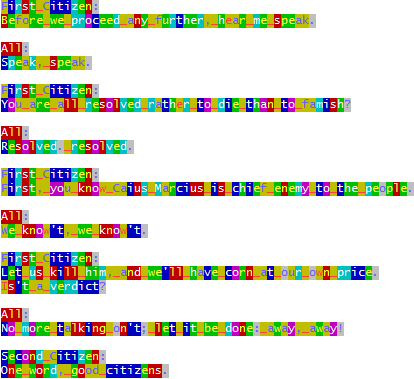}
\caption{Hybrid LSTM component: colors correspond to 10 k-means clusters on hidden state vectors. Yellow cluster (with red font) identifies spaces. Grey cluster identifies punctuation (except commas). Purple cluster finds some 'y' and 'o' letters.}
\label{subfig:shak_lstm}
\end{subfigure}
\caption{Visualizing HMM and LSTM states on Shakespeare data for the hybrid with 10 LSTM state dimensions and 10 HMM states. The HMM and LSTM components learn some complementary features in the text: while both learn to identify spaces, the LSTM does not completely identify punctuation or pick up on vowels, which the HMM has already done.\label{fig:viz_hybrid_shak}}
\begin{subfigure}{0.49\textwidth}
\centering
\includegraphics[scale=.57]{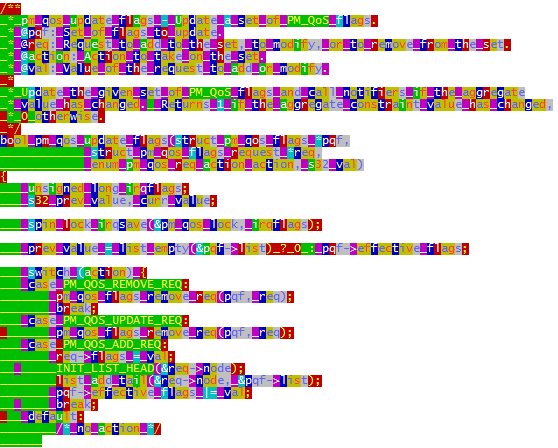}
\caption{Hybrid HMM component: colors correspond to 10 HMM states. Distinguishes comments and indentation spaces (green with yellow font) from other spaces (purple). Red cluster (with yellow font) identifies punctuation and brackets. Green cluster (yellow font) also finds capitalized variable names.}
\label{subfig:linux_hmm}
\end{subfigure}\hfill
\begin{subfigure}{0.49\textwidth}
\centering
\includegraphics[scale=.57]{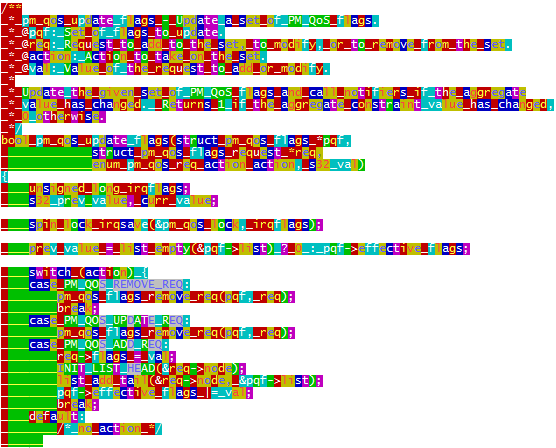}
\caption{Hybrid LSTM component: colors correspond to 10 k-means clusters on hidden state vectors. Distinguishes comments, spaces at beginnings of lines, and spaces between words (red with white font) from indentation spaces (green with yellow font). Opening brackets are red (yellow font) and closing brackets are green (white font).} \label{subfig:linux_lstm}
\end{subfigure}
\caption{Visualizing HMM and LSTM states on Linux data for the hybrid with 10 LSTM state dimensions and 10 HMM states. The HMM and LSTM components learn some complementary features in the text related to spaces and comments. \label{fig:viz_hybrid_linux}}
\end{figure*}

We interpret the HMM and LSTM states in the hybrid algorithm with 10 LSTM state dimensions and 10 HMM states in Figures \ref{fig:viz_hybrid_shak} and \ref{fig:viz_hybrid_linux}, showing which features are identified by the HMM and LSTM components. In Figures \ref{subfig:shak_hmm} and \ref{subfig:linux_hmm}, we color-code the training data with the 10 HMM states. In Figures \ref{subfig:shak_lstm} and \ref{subfig:linux_lstm}, we apply k-means clustering to the LSTM state vectors, and color-code the training data with the clusters. The HMM and LSTM states pick up on spaces, indentation, and special characters in the data (such as comment symbols in Linux data). We see some examples where the HMM and LSTM complement each other, such as learning different things about spaces and comments on Linux data, or punctuation on the Shakespeare data. In Figure \ref{fig:decision_trees}, we see that some individual LSTM hidden state dimensions identify similar features, such as comment symbols in the Linux data.

\begin{table}
\small
\caption{Predictive loglikelihood comparison on the text data sets (sorted by validation set performance).}
\label{tab:results}
\begin{tabular}{l|l|lll|ll}
\multicolumn{1}{c}{\rotatebox{90}{Data}} & \multicolumn{1}{c}{\rotatebox{90}{Method}} & \multicolumn{1}{c}{\rotatebox{90}{Parameters}} & \multicolumn{1}{c}{\rotatebox{90}{LSTM dims}} & \multicolumn{1}{c}{\rotatebox{90}{HMM states}} & \multicolumn{1}{c}{\rotatebox{90}{Validation}} & \multicolumn{1}{c}{\rotatebox{90}{Training}} \\\hline
\multirow{19}{*}{\rotatebox{90}{Shakespeare}}
       & Continuous HMM & 1300 &     & 20 & -2.74 & -2.75 \\
       & Discrete HMM   & 650  &     & 10 & -2.69 & -2.68 \\
       & Discrete HMM   & 1300 &     & 20 & -2.5  & -2.49 \\
       & LSTM           & 865  & 5   &    & -2.41 & -2.35 \\
       & Hybrid         & 1515 & 5   & 10 & -2.3  & -2.26 \\
       & Hybrid         & 2165 & 5   & 20 & -2.26 & -2.18 \\
       & LSTM           & 2130 & 10  &    & -2.23 & -2.12 \\
       & Joint hybrid   & 1515 & 5   & 10 & -2.21 & -2.18 \\
       & Hybrid         & 2780 & 10  & 10 & -2.19 & -2.08 \\
       & Hybrid         & 3430 & 10  & 20 & -2.16 & -2.04 \\
       & Hybrid         & 4445 & 15  & 10 & -2.13 & -1.95 \\
       & Joint hybrid   & 3430 & 10  & 10 & -2.12 & -2.07 \\
       & LSTM           & 3795 & 15  &    & -2.1  & -1.95 \\
       & Hybrid         & 5095 & 15  & 20 & -2.07 & -1.92 \\
       & Hybrid         & 6510 & 20  & 10 & -2.05 & -1.87 \\
       & Joint hybrid   & 4445 & 15  & 10 & -2.03 & -1.97 \\
       & LSTM           & 5860 & 20  &    & -2.03 & -1.83 \\
       & Hybrid         & 7160 & 20  & 20 & -2.02 & -1.85 \\
       & Joint hybrid   & 7160 & 20  & 10 & -1.97 & -1.88 \\
      \hline
\multirow{18}{*}{\rotatebox{90}{Linux Kernel}}
       & Discrete HMM   & 1000 &     & 10 & -2.76 & -2.7  \\
       & Discrete HMM   & 2000 &     & 20 & -2.55 & -2.5  \\
       & LSTM           & 1215 & 5   &    & -2.54 & -2.48 \\
       & Joint hybrid   & 2215 & 5   & 10 & -2.35 & -2.26 \\
       & Hybrid         & 2215 & 5   & 10 & -2.33 & -2.26 \\
       & Hybrid         & 3215 & 5   & 20 & -2.25 & -2.16 \\
       & Joint hybrid   & 4830 & 10  & 10 & -2.18 & -2.08 \\
       & LSTM           & 2830 & 10  &    & -2.17 & -2.07 \\
       & Hybrid         & 3830 & 10  & 10 & -2.14 & -2.05 \\
       & Hybrid         & 4830 & 10  & 20 & -2.07 & -1.97 \\
       & LSTM           & 4845 & 15  &    & -2.03 & -1.9  \\
       & Joint hybrid   & 5845 & 15  & 10 & -2.00 & -1.88 \\
       & Hybrid         & 5845 & 15  & 10 & -1.96 & -1.84 \\
       & Hybrid         & 6845 & 15  & 20 & -1.96 & -1.83 \\
       & Joint hybrid   & 9260 & 20  & 10 & -1.90 & -1.76 \\
       & LSTM           & 7260 & 20  &    & -1.88 & -1.73 \\
       & Hybrid         & 8260 & 20  & 10 & -1.87 & -1.73 \\
       & Hybrid         & 9260 & 20  & 20 & -1.85 & -1.71 \\
\hline    
\multirow{18}{*}{\rotatebox{90}{Penn Tree Bank}}
       & Continuous HMM & 1000 & 100 & 20 & -2.58 & -2.58 \\
       & Discrete HMM   & 500  &     & 10 & -2.43 & -2.43 \\
       & Discrete HMM   & 1000 &     & 20 & -2.28 & -2.28 \\
       & LSTM           & 715  & 5   &    & -2.22 & -2.22 \\
       & Hybrid         & 1215 & 5   & 10 & -2.14 & -2.15 \\
       & Joint hybrid   & 1215 & 5   & 10 & -2.08 & -2.08 \\
       & Hybrid         & 1715 & 5   & 20 & -2.06 & -2.07 \\
       & LSTM           & 1830 & 10  &    & -1.99 & -1.99 \\
       & Hybrid         & 2330 & 10  & 10 & -1.94 & -1.95 \\
       & Joint hybrid   & 2830 & 10  & 10 & -1.94 & -1.95 \\
       & Hybrid         & 2830 & 10  & 20 & -1.93 & -1.94 \\
       & LSTM           & 3345 & 15  &    & -1.82 & -1.83 \\
       & Hybrid         & 3845 & 15  & 10 & -1.81 & -1.82 \\
       & Hybrid         & 4345 & 15  & 20 & -1.8  & -1.81 \\
       & Joint hybrid   & 6260 & 20  & 10 & -1.73 & -1.74 \\
       & LSTM           & 5260 & 20  &    & -1.72 & -1.73 \\
       & Hybrid         & 5760 & 20  & 10 & -1.72 & -1.72 \\
       & Hybrid         & 6260 & 20  & 20 & -1.71 & -1.71 \\
      \hline
\end{tabular}
\end{table}

\section{Conclusion and future work}
Hybrid HMM-RNN approaches combine the interpretability of HMMs with the predictive power of RNNs. Sometimes, a small hybrid model can perform better than a standalone LSTM of the same size. We use visualizations to show how the LSTM and HMM components of the hybrid algorithm complement each other in terms of features learned in the data.

\bibliography{int_bibliography}
\bibliographystyle{icml2016}

\end{document}